\documentclass[lettersize,journal]{IEEEtran}
\usepackage{amsmath,amsfonts}
\usepackage{algorithmic}
\usepackage{algorithm}
\usepackage{array}
\usepackage[caption=false,font=normalsize,labelfont=sf,textfont=sf]{subfig}
\usepackage{textcomp}
\usepackage{stfloats}
\usepackage{url}
\usepackage{verbatim}
\usepackage{graphicx}
\usepackage{cite}
\usepackage{multirow}
\usepackage{makecell}
\usepackage{array}

\begin{document}
\title{Offline–Online Hierarchical 3D Global Relocalization With Synthetic LiDAR Sensing and Descriptor-Space Retrieval}

\author{Jiahua Ren, Kai Shen, Muhua Zhang, and Lei Ma
\thanks{This work was supported by the National Natural Science Foundation of China (Grants 62203371). (Corresponding author: Kai Shen). }
\thanks{Jiahua Ren, Kai Shen, Muhua Zhang and Lei Ma are with the School of Electrical Engineering, Southwest Jiaotong University, Chengdu 611756, China (e-mail: shenkai@swjtu.edu.cn).
}
}

\maketitle

\begin{abstract}

3D global relocalization is one of the key capabilities for mobile robots in practical applications. However, in large-scale spaces, existing methods often suffer from prolonged online relocalization time due to factors such as the massive pose search space and high computational overhead. To address these issues, this paper proposes an offline-online hierarchical framework that decouples the search space. In the offline phase, candidate positions and their corresponding geometric descriptor indices are generated in the map by simulating LiDAR scans within the grid map. In the online phase, a coarse pose estimate is first obtained via global retrieval, followed by point cloud registration to output precise 6-DoF pose estimates. Real-world experiments demonstrate that the proposed method achieves an average relocalization time of 3 s and an average localization accuracy of 8~cm in 3D environments. Compared with existing global relocalization methods, the proposed method achieves an order-of-magnitude improvement in computational efficiency while delivering comparable relocalization accuracy.
\end{abstract}

\begin{IEEEkeywords}
    3D Global Relocalization, LiDAR Scan, 6-DoF Pose Estimation, Uniform Sampling
\end{IEEEkeywords}

\section{Introduction}                   
\IEEEPARstart{I}{n} scenarios such as warehousing logistics and industrial inspection, robots typically rely on pre-constructed two-dimensional (2D) or three-dimensional (3D) maps to execute tasks. When the initial pose is unknown (the kidnapped robot problem, KRP), or when localization is lost due to aggressive motion, severe occlusions, or degraded sensing conditions, a task-executing robot must promptly recover its pose on the prior map based on current LiDAR observations to ensure operational safety and task continuity. However, although many methods have been proposed for 3D global relocalization \cite{meng2021efficient}, \cite{shi2024fast}, \cite{su2017global}, \cite{chen2024fusednet}, achieving both real-time performance and high accuracy in large-scale 3D environments remains highly challenging, primarily due to the need for efficient processing and matching of high-dimensional LiDAR observations.

For 3D global relocalization, the robot must search for global 6-DoF candidate poses on a 3D map. The number of candidates grows combinatorially with the map size and the search resolution, which significantly raises the difficulty of relocalization. Even classical global registration methods (e.g., GO-ICP \cite{yang2015go}) may fail to align in large-scale scenes due to repetitive structures, outliers, and limited overlap, and they often incur high computational cost. In practice, many systems therefore rely on a coarse initial pose provided manually; however, in large environments such a coarse estimate is often difficult to obtain accurately, which limits the applicability of these local relocalization schemes. On the other hand, the data scale and computational burden in 3D environments are also major limiting factors. Whether using vision-based feature extraction and matching or LiDAR-based point-cloud processing and registration, large-scale scenarios require handling high-dimensional and dense prior maps simultaneously with real-time observations. This leads to substantial computational demand, making full-map matching strategies difficult to deploy directly on robots with limited payload and onboard computing resources. Consequently, in practical applications, existing global relocalization methods often struggle to achieve both real-time performance and high accuracy in large environments: more exhaustive global search and stricter verification typically introduce higher computational latency, whereas reducing computation can degrade accuracy and robustness.

\begin{figure}[t]
    \centering
    \includegraphics[width=0.8\linewidth]{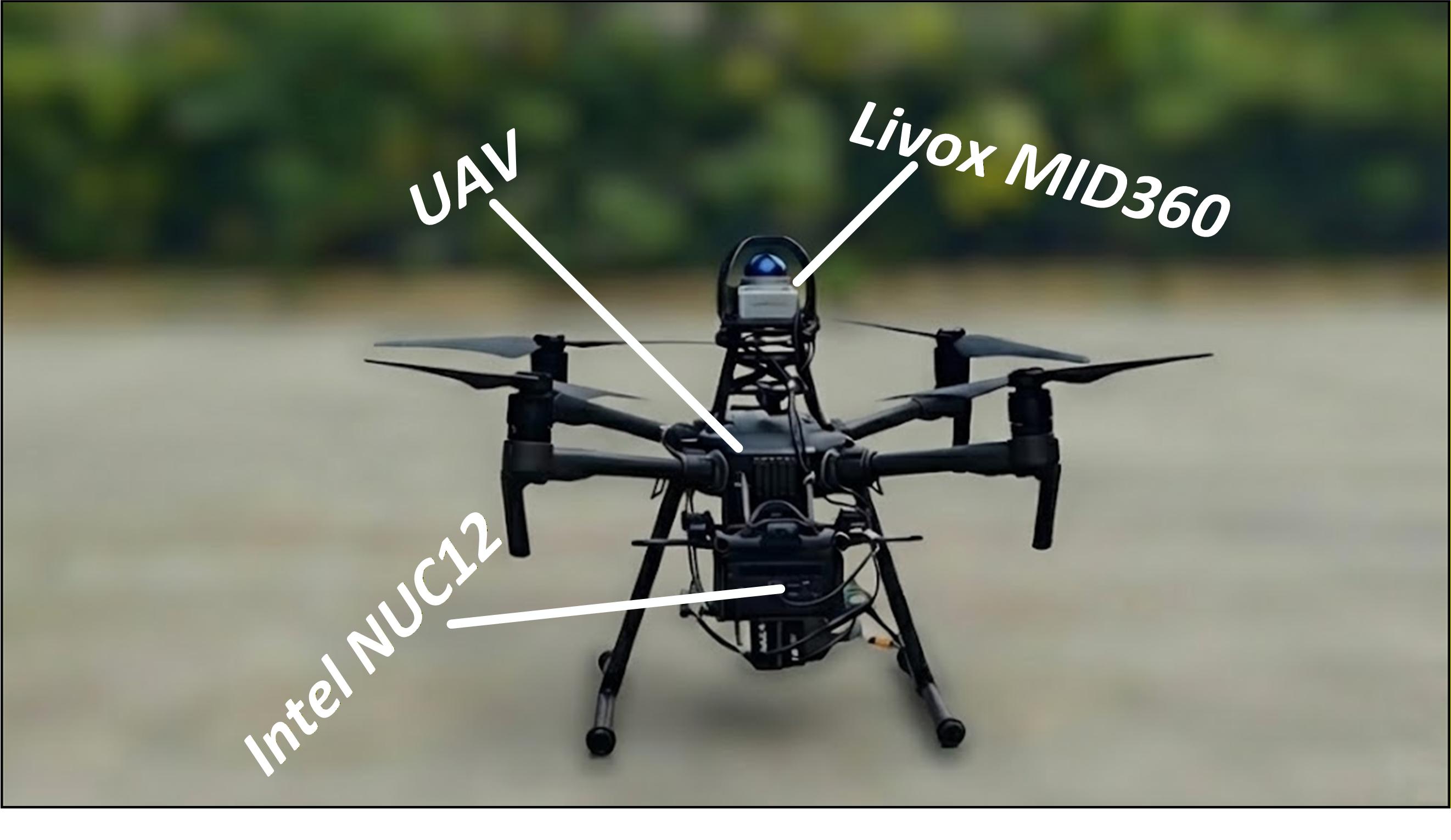}
    \caption{\textbf{Experimental UAV platform.} A Livox MID360 LiDAR is mounted on top of the UAV for 3D perception, while an Intel NUC12 onboard computer handles real-time mapping and relocalization.}
    \label{fig:UAV}
\end{figure}

To address the above challenges, we propose a 3D LiDAR-based global relocalization framework for mobile robots. The method supports pose initialization from arbitrary locations in a prior map and is applicable to platforms with different motion dimensionalities. Our pipeline first uses a rotation-invariant Scan Context descriptor \cite{kim2018scan} to retrieve candidate regions and estimate yaw, then utilizes default roll and pitch to form a 6-DoF input, which is fed into point-cloud registration to refine the pose. Experiments show that, given only online LiDAR scans and a prior map, the proposed method achieves centimeter-level positional error and degree-level orientation error with runtimes on the order of seconds in large-scale outdoor environments.

In addition, we combine 3D pose sampling with ray-casting-based synthetic LiDAR scans and global matching to generate global pose hypotheses over a prior occupancy grid map, and to precompute a corresponding point cloud descriptor for each hypothesis. To validate the proposed approach, we implement the system on an aerial platform equipped with a Livox MID360 LiDAR and an Intel NUC12 onboard computer, as shown in Fig.~\ref{fig:UAV}.

The contributions of this work are as follows:
\begin{itemize}
\item \textbf{3D uniform sampling method:} In a 3D occupancy grid, a uniform sampling method is designed by extending 3D RRT with the Clearance Constraint, Minimum-Separation Constraint, Geometric Observability Check, and Sampling Termination Strategy to generate uniformly distributed feasible candidate positions.
\item \textbf{Hierarchical pose-matching method using synthetic scans and descriptors:} In the offline stage, a virtual LiDAR is used to scan the occupancy grid map at sampled positions and construct a map-side point-cloud descriptor database; in the online stage, real-time LiDAR scans are converted into query descriptors in a descriptor-consistent manner for global pose matching, followed by GN-ICP-based pose refinement.
\item \textbf{Global relocalization framework:} Unlike other relocalization algorithms, the proposed method employs a novel integration strategy that achieves global relocalization in large-scale environments using only real-time LiDAR scans and a prior map.
\end{itemize}

\section{Related Work}                  
Global relocalization, which aims to estimate a robot's pose in a prior map without any initial pose estimate, has been extensively studied. In mobile robotics, existing global relocalization approaches can be broadly categorized by sensor modality into vision-based methods \cite{glocker2013real}, \cite{schneider2018maplab} and LiDAR-based methods \cite{zhang20223d}, \cite{zhao2020indoor}, \cite{vaquero2019improving}.

In visual relocalization, many methods build on local feature extraction and matching to establish data association, using hand-crafted descriptors such as SIFT \cite{lowe1999object}, ORB \cite{rublee2011orb}, and BRIEF \cite{calonder2010brief}, as well as learned matchers like SuperGlue \cite{sarlin2020superglue}. Feature-based relocalization has been extensively studied; for instance, \cite{se2002global} performs global relocalization with SIFT, and VINS-Mono \cite{qin2018vins} uses BRIEF features within a visual-inertial pipeline for localization and relocalization. Beyond local matching, several works leverage prior maps with feature constraints to achieve high-precision global relocalization. Hao et al. \cite{hao2023global} construct a VIO-based feature map and localize by associating online observations with map features, while Qin et al. \cite{qin2018vins} further introduces global pose-graph optimization to fuse local VIO with prebuilt maps. In the vision-only setting, Campos et al. \cite{campos2021orb} propose an Atlas mechanism for relocalization and map merging across multiple maps. To improve scalability, Sattler et al. \cite{sattler2016efficient} propose prioritized 2D--3D matching for efficient localization in large point-cloud maps. Topological and semantic-topological methods have also been explored to reduce the dependence on precise metric information \cite{liu2019global,angeli2009visual}. Earlier vision-based studies also considered landmark-assisted localization. Hashimoto et al.\cite{hashimoto1999mobile} used color signboards to support absolute and relative localization for indoor mobile robots.

\begin{figure*}[h]
    \centering
    \includegraphics[width=1.0\linewidth]{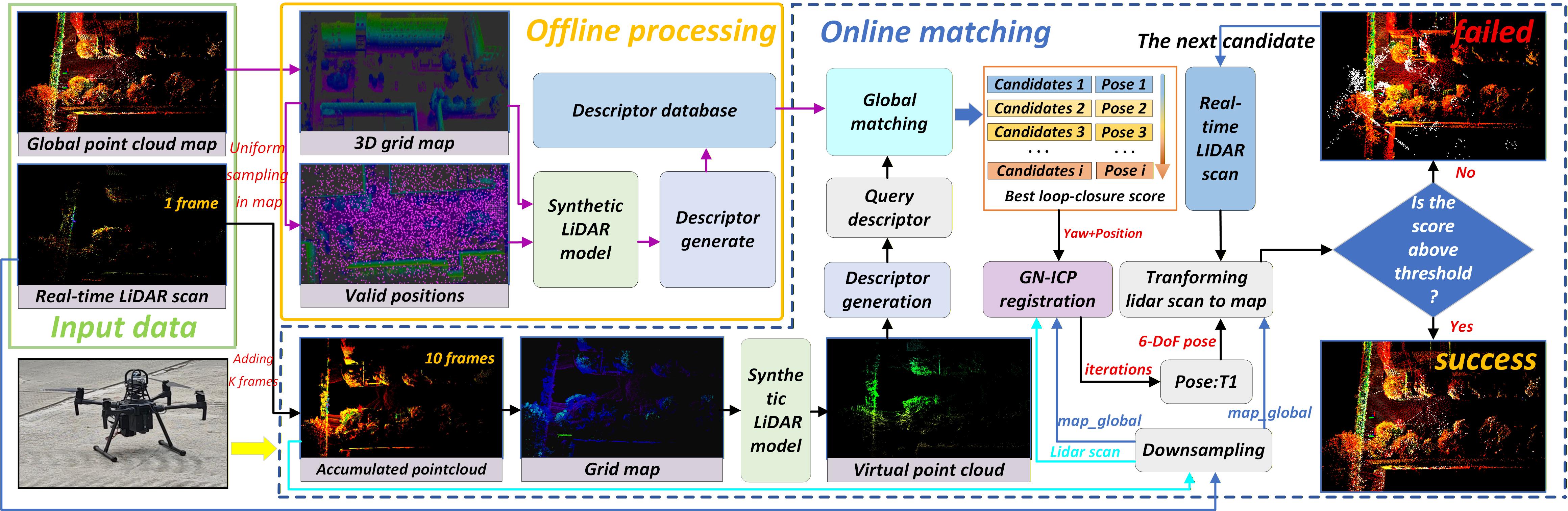}
    \caption{\textbf{Overall pipeline of the proposed LiDAR-based global 3D relocalization framework.} The system takes as input a prior global point cloud map and real-time LiDAR scans. In the offline processing stage (orange), valid robot poses are uniformly sampled in the prior 3D occupancy grid map, a virtual LiDAR model is used to generate synthetic scans, and a descriptor database is constructed from the resulting virtual point clouds. In the online matching stage (blue), \textbf{$K_f$} consecutive LiDAR frames are accumulated to build a local grid map, a query descriptor is generated and matched against the database for global matching, and the best candidate pose is refined via \textbf{GN-ICP} registration on a downsampled point cloud to obtain the final \textbf{6-DoF} relocalization result.}
    \label{fig:system}
\end{figure*}

Compared with cameras, LiDAR provides a wider field of view and a longer sensing range, and is less affected by illumination and other environmental conditions, which makes them more suitable for large-scale environments \cite{zhou2023comparative}. Because of the large amount of data in 3D point clouds, directly processing raw point clouds is very resource-intensive. Therefore, many studies \cite{ref21}, \cite{ref22}, \cite{ref23}, \cite{ref27} employ point cloud descriptors to represent the characteristics of 3D point clouds, and then perform subsequent processing based on these descriptors.

Many works follow a similar idea by transforming raw point clouds into more tractable 2D/2.5D representations for place recognition and loop closure. For example, LiDAR Iris converts a single LiDAR scan into a LiDAR-Iris image and derives a binarized signature via LoG-Gabor filtering and thresholding for efficient matching \cite{wang2020lidar}.
Under a planar-motion assumption, HOPN \cite{luo2022lidar} and SPI \cite{li2021lidar} project point clouds into bird's-eye-view images and perform matching using statistical descriptors. In our relocalization setting, pose estimation must efficiently retrieve plausible candidates from a large map and provide reliable initialization for subsequent fine registration. We therefore adopt Scan Context \cite{kim2018scan} as the global descriptor and retrieval module. It encodes each scan as a 2.5D polar-grid matrix and performs yaw alignment through circular shifts. In addition, a lightweight ring-key-based indexing scheme implemented with a KD-tree enables fast candidate retrieval and reduces the online retrieval cost.

Beyond descriptor-driven candidate retrieval, another line of work focuses on exploiting geometric consistency in point clouds for direct pose estimation and match verification. Representative direct registration methods include GO-ICP \cite{yang2015go} and NDT \cite{biber2003normal}. For structured indoor scenes, Yao et al. \cite{yao2024crtf} further proposed a coarse-to-fine reflector-based LiDAR positioning framework, in which coarse alignment is followed by ICP-based refinement to improve robustness.  

In addition, under a planar 3-DoF motion assumption, several studies improve large-area global localization efficiency via multiresolution representations and structured search. For example, Meng et al. \cite{meng2021efficient} perform coarse-to-fine global pose search using multiscale maps together with an optimized branch-and-bound strategy. For 3D LiDAR global relocalization, existing studies have explored BEV-projection-based matching and Siamese-network-based global embedding learning \cite{luo2025bevplace++,yin20193d}. Wang et al. \cite{wang2021pointloc} proposed PointLoc, an end-to-end LiDAR relocalization method that directly regresses the 6-DoF pose from a single point cloud. Cross-modal approaches have also been explored. For example, Cattaneo et al. \cite{cattaneo2020global} learned a shared embedding space between 2D images and 3D LiDAR maps for vision-based relocalization. 

However, many existing methods still face challenges in computational cost and scalability, making it difficult to achieve both high relocalization accuracy and online efficiency in large-scale environments. Meanwhile, some 2D-representation-based localization strategies are efficient, but the loss of geometric information caused by dimensionality reduction often limits them to planar-motion assumptions, which do not readily extend to scenes with complex 3D structures or noticeable vertical motion. To address these issues, this paper proposes a framework that moves computationally expensive sampling and descriptor construction to an offline stage. During online operation, it performs efficient candidate global matching in descriptor space and refines the pose using GN-ICP, enabling accurate and robust global 3D relocalization in large outdoor environments with only a single LiDAR.

\section{Methodology}                  
\subsection{System Overview}
In GNSS-denied or unreliable-GNSS conditions, robots can only rely on onboard perception to estimate their pose with respect to a prior map, which makes global relocalization challenging. To address this problem, a 3D global relocalization framework is proposed. As illustrated in Fig. \ref{fig:system}, the framework comprises two core modules: an offline processing phase and an online processing phase.

In the offline processing stage, the proposed method takes a prior global point cloud map as input and builds a descriptor database for subsequent online matching. First, the point cloud is converted into a 3D occupancy grid map, which keeps the basic scene structure and avoids handling dense point clouds in large-scale environments, reducing the cost of subsequent position sampling and synthetic scan generation. Second, inspired by \cite{zhang2025tackling}, we propose a 3D RRT-based multi-hypothesis uniform sampling approach to generate feasible robot position candidates on the occupancy grid. Third, we design a virtual LiDAR model possessing the same field of view and angular resolution as the real LiDAR. This model performs ray casting on the grid map at each sampled position (with a fixed LiDAR orientation) to generate the corresponding synthetic point cloud. Finally, each synthetic point cloud is transformed into the virtual LiDAR frame and encoded as a Scan-Context-style \cite{kim2018scan} descriptor. For each sampled position, the descriptor generated from the synthetic point cloud is stored in the global database along with its corresponding 3D coordinates, which later serves as the prior search space for online 3D global relocalization.

During the online matching stage, the system queries the offline descriptor database using current LiDAR descriptors to retrieve potential candidates, which are subsequently refined via geometric verification to compute the precise relocalization result. Given that the utilized LiDAR operates in a non-repetitive scanning mode, $K_f$ consecutive frames are first accumulated as the current input, where $K_f$ is typically set to 10--20. This point cloud is voxelized into a local 3D grid map, on which a virtual LiDAR performs a synthetic scan to generate a virtual point cloud; a descriptor of the current environment is then computed from this virtual point cloud using the same encoding as in the offline stage. Global matching is performed to identify the top $K_c$ candidates with the smallest matching scores, where $K_c=5$ in our experiments. By combining the 3D position retrieved from the database with the yaw angle estimated via descriptor alignment, a 4-DoF initial pose hypothesis $(x, y, z, \psi)$ is constructed. It is then augmented with default roll and pitch to obtain a full 6-DoF initial pose for subsequent registration. The global map and the current accumulated point cloud are both downsampled, and each candidate pose is taken as the initial estimate for ICP registration. ICP is then executed to refine the candidate pose and compute a matching error score. If this score falls below a predefined threshold, the corresponding pose is accepted as the relocalization result; otherwise, the result is discarded and ICP is run again starting from the next candidate pose.

By offloading complex and computationally intensive data processing tasks to the offline stage, and performing efficient descriptor-based global matching followed by GN-ICP refinement online, the proposed framework enables robust global 3D relocalization using only a single LiDAR. Moreover, when the environment undergoes non-disruptive changes (such as local structural modifications or varying occlusions), the proposed framework still maintains a high relocalization success rate and localization accuracy.

\subsection{Uniform Sampling}
This section presents a goal-free, constraint-aware sampling strategy to generate a global set of candidate positions on the 3D occupancy grid map for offline descriptor database construction.
\subsubsection{Problem Formulation}
A binary occupancy grid $\mathcal{M}$ with resolution $r$ is used, where $\chi(\mathbf{x})\in\{0,1\}$ is defined such that $\chi(\mathbf{x})=1$ for occupied voxels and $\chi(\mathbf{x})=0$ for free voxels. The feasible sampling set is then $\mathcal{F}=\{\mathbf{x}\in\Omega\mid \chi(\mathbf{x})=0\}$, where $\Omega\subset\mathbb{R}^3$ is the map bounding box. In the offline stage, we do not specify a goal; rather, we generate a global set of candidate \emph{positions} $\mathcal{S}=\{\mathbf{p}_i\}_{i=1}^{N_p}$ (where $N_p=|\mathcal{S}|$) for subsequent virtual LiDAR scanning and descriptor database construction.

\begin{figure}
    \centering
    \includegraphics[width=1.0\linewidth]{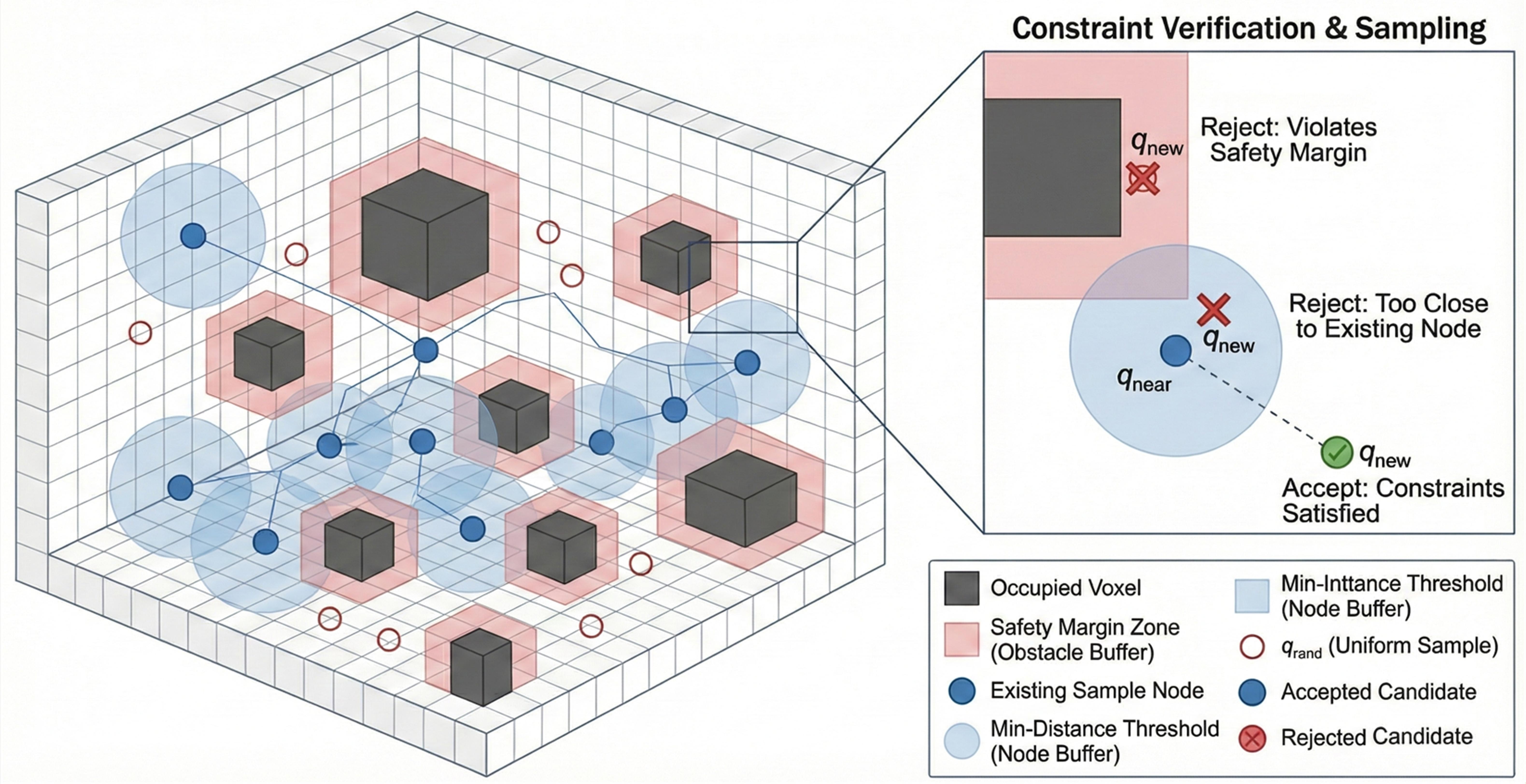}
    \caption{Constraint-aware uniform sampling in a 3D occupancy grid map.}
    \label{RRT_sim}
\end{figure}

\begin{figure}
    \centering
    \includegraphics[width=1.0\linewidth]{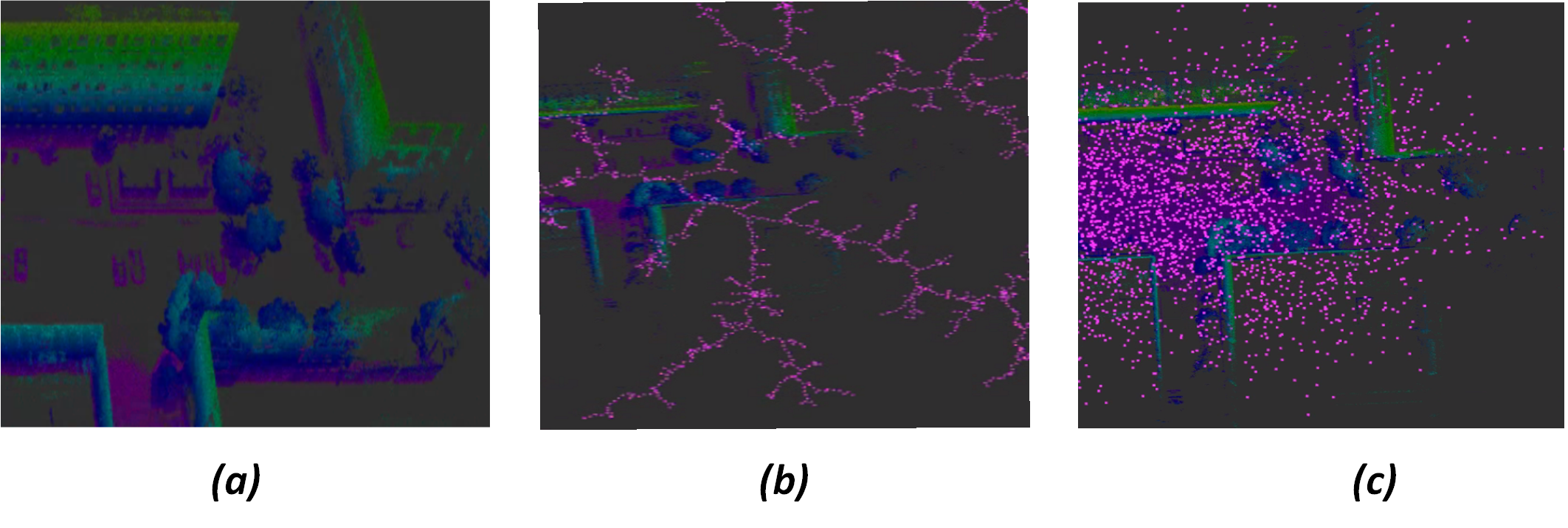}
    \caption{Sampling on the 3D occupancy grid map with the same number of sampling steps: (a) 3D occupancy grid map; (b) standard 3D RRT sampling without the proposed constraints; (c) proposed constraint-aware RRT-based uniform sampling using the same number of samples as in (b).}
    \label{fig:RRT_real}
\end{figure}

\subsubsection{Constraint-Aware Acceptance Criterion}
To reduce redundant samples, improve spatial coverage uniformity, and filter out positions with insufficient geometric support, three constraints are introduced for each candidate point $\mathbf{p}_{\mathrm{new}}$, as illustrated in Fig.~\ref{RRT_sim}. A candidate is accepted only if
\begin{equation}
\begin{aligned}
\operatorname{Accept}\!\left(\mathbf{p}_{\mathrm{new}}\right)
&= \mathcal{C}_{\mathrm{clr}}\!\left(\mathbf{p}_{\mathrm{new}}\right)
\wedge \mathcal{C}_{\mathrm{sep}}\!\left(\mathbf{p}_{\mathrm{new}},\mathcal{S}\right)\\
&\quad \wedge \mathcal{C}_{\mathrm{ray}}\!\left(\mathbf{p}_{\mathrm{new}}\right).
\end{aligned}
\label{eq:accept}
\end{equation}

\textbf{Clearance constraint:}
Let $r_{\mathrm{uav}}$ denote the effective UAV radius and $r_{\mathrm{safe}}$ an additional safety margin. The minimum required clearance is defined as $r_{\mathrm{clr}}=r_{\mathrm{uav}}+r_{\mathrm{safe}}$. Specifically, the following condition must be satisfied for a candidate position $\mathbf{p}$:
\begin{equation}
\min_{\mathbf{y}\in\mathcal{V}_{\mathrm{occ}}}\|\mathbf{p}-\mathbf{y}\|\ge r_{\mathrm{clr}}.
\label{eq:clearance}
\end{equation}
This constraint is equivalent to inflating obstacles by $r_{\mathrm{clr}}$ in configuration space, which suppresses dense sampling near obstacles.

\textbf{Minimum-separation constraint:}
To avoid overly dense sampling, a minimum inter-sample distance $r_{\mathrm{sep}}$ (e.g., 1.2~m) is imposed. Specifically, a candidate position $\mathbf{p}$ must satisfy
\begin{equation}
\min_{\mathbf{p}_i\in\mathcal{S}}\|\mathbf{p}-\mathbf{p}_i\|\ge r_{\mathrm{sep}}.
\label{eq:minimum_separation}
\end{equation}
This can be interpreted as a Poisson-disk (blue-noise) spacing rule that promotes more uniform coverage of the free space.

\textbf{Geometric Observability Check:}
To prevent samples from being placed in geometrically uninformative regions, we cast rays from a candidate position $\mathbf{p}$ along a set of directions $\mathcal{D}=\{\mathbf{d}_\ell\}_{\ell=1}^{K_r}$ with a maximum range $R$, where $K_r$ denotes the number of ray directions, and count how many directions hit an occupied voxel. Specifically, we define a hit indicator $h(\mathbf{p},\mathbf{d}_\ell)\in\{0,1\}$, where $h(\mathbf{p},\mathbf{d}_\ell)=1$ if the ray from $\mathbf{p}$ along $\mathbf{d}_\ell$ hits an occupied voxel within range $R$, and $h(\mathbf{p},\mathbf{d}_\ell)=0$ otherwise. We require the number of hit directions to satisfy
\begin{equation}
H(\mathbf{p})=\sum_{\ell=1}^{K_r} h(\mathbf{p},\mathbf{d}_\ell)\ \ge\ N_{\mathrm{hit}},
\label{eq:hit_count}
\end{equation}
where $N_{\mathrm{hit}}$ is a threshold (e.g., $N_{\mathrm{hit}}=2$), which filters out positions that lack sufficient observable structure for reliable descriptor matching.

\subsubsection{Sampling Termination Strategy}
To avoid over- or under-sampling across environments of different scales, we adopt a window-based early stopping strategy based on the yield of valid samples. Specifically, we group iterations into windows of length $W$ (e.g., $W=10000$) and denote by $s_t$ the number of newly accepted valid positions in the $t$-th window. 
The reference baseline and the stopping criterion are defined as
\begin{equation}
\left\{
\begin{aligned}
\mu_{\mathrm{ref}} &= \frac{1}{N_{\mathrm{ref}}}\sum_{t=t_s}^{t_e} s_t,\\
N_{\mathrm{ref}} &= t_e - t_s + 1,\\
s_t &< \alpha\,\mu_{\mathrm{ref}},
\end{aligned}
\right.
\label{eq:early_stop}
\end{equation}
where $t_s$ and $t_e$ denote the start and end indices of the reference window range, respectively. Here, $\mu_{\mathrm{ref}}$ denotes the average number of newly accepted valid positions per window over the reference window range, and $\alpha\in(0,1)$ is a threshold ratio controlling the early-stopping sensitivity. In our experiments, we set $t_s=5$, $t_e=10$, and $\alpha=0.4$.

Overall, the proposed constraint-aware sampling yields a compact and spatially well-distributed candidate set $\mathcal{S}$ by jointly enforcing clearance, minimum separation, and a geometric observability check. The effect of the proposed constraints on the spatial distribution of samples is illustrated in Fig.~\ref{fig:RRT_real}. Sampling is terminated via a windowed saturation-aware criterion, and the resulting candidates are used to generate synthetic scans and build the offline descriptor database.

\subsection{Virtual LiDAR Scanning and Descriptor Construction}

\begin{figure}
    \centering
    \includegraphics[width=1.0\linewidth]{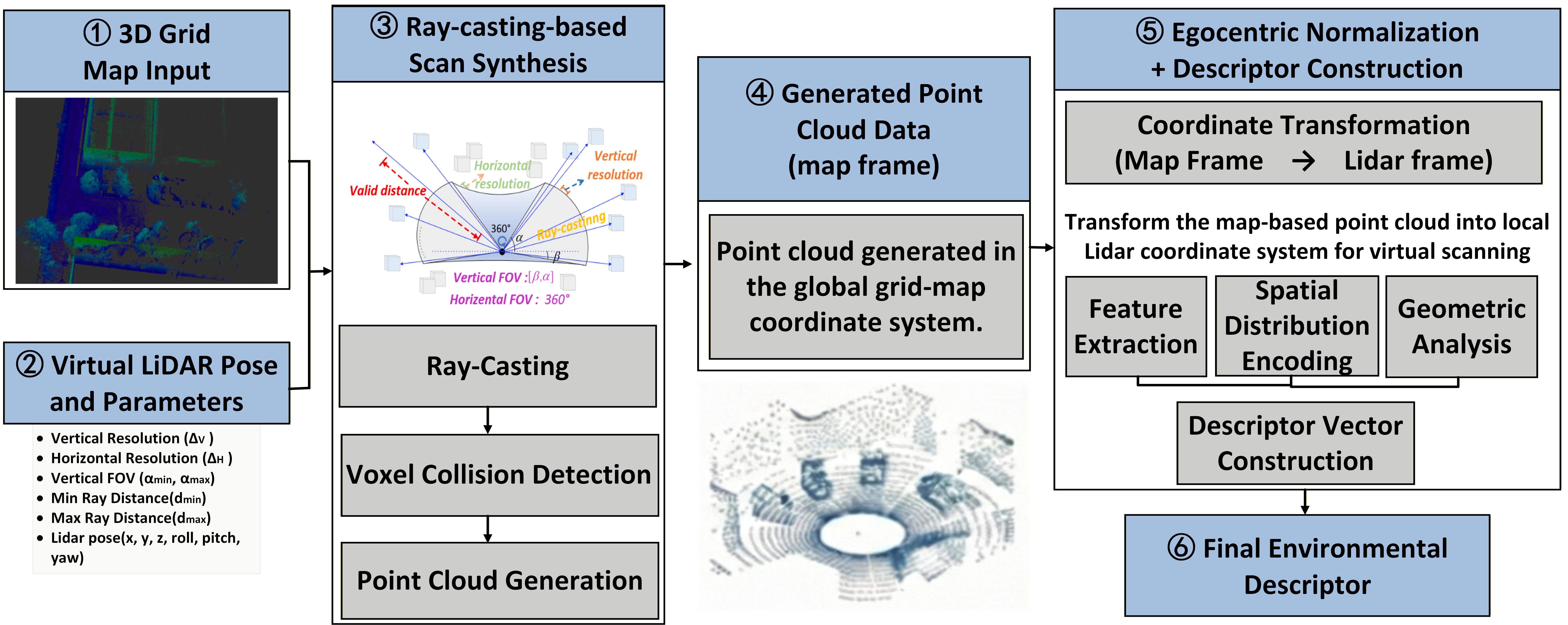}
    \caption{Pipeline of ray-casting-based scan synthesis and egocentric descriptor construction.}
    \label{fig:lidar_visual}
\end{figure}

For each sampled position $\mathbf{p}_i \in \mathcal{S}$ on the 3D occupancy grid map $\mathcal{M}$, we instantiate a virtual LiDAR that \emph{matches the real sensor's field-of-view and angular sampling pattern} (azimuth/elevation discretization) and perform \emph{beam-based ray casting} on $\mathcal{M}$. Then, \emph{first-return} ray casting is performed: for each beam direction $\mathbf{d}_\ell$ in $\mathcal{D}=\{\mathbf{d}_\ell\}_{\ell=1}^{K_r}$, a ray is traced from $\mathbf{p}_i$ up to range $R$ and stops at the first occupied voxel ($\chi(\cdot)=1$). The corresponding hit point is recorded as a return, while rays with no hit within $R$ are discarded. Collecting all returns forms the synthesized scan $\mathcal{P}_i^{\mathcal{M}}$ in the map frame. The specific process is illustrated in Fig.~\ref{fig:lidar_visual}.

Since the virtual LiDAR uses a \emph{fixed LiDAR orientation} for all synthesized scans, we set a constant rotation $\mathbf{R}_0$ and define the virtual LiDAR pose at $\mathbf{p}_i$ as $\mathbf{T}_i=\left[\mathbf{R}_0 \,\middle|\, \mathbf{p}_i\right]$. For any synthesized point $\mathbf{x}\in \mathcal{P}_i^{\mathcal{M}}$ (map frame), its coordinates in the LiDAR frame are
\begin{equation}
\mathbf{x}^{\mathcal{L}}
= \mathbf{R}_0^{\top}\!\left(\mathbf{x}-\mathbf{p}_i\right),
\qquad
\mathcal{P}_i^{\mathcal{L}}
= \left\{\mathbf{x}^{\mathcal{L}} \,\middle|\, \mathbf{x}\in \mathcal{P}_i^{\mathcal{M}}\right\}.
\label{eq:map_to_lidar}
\end{equation}

Following Scan Context \cite{kim2018scan}, we partition $\mathcal{P}_i^{\mathcal{L}}$ on the $xy$-plane into $N_r$ \emph{rings} and $N_s$ \emph{sectors} up to range $L_{\max}$, and denote the point set in bin $(u,v)$ by $\mathcal{B}_{u,v}$. The Scan Context matrix $\mathbf{I}_i \in \mathbb{R}^{N_r\times N_s}$ is then defined by \emph{max-height encoding} (empty bins set to $0$) as
\begin{equation}
\mathbf{I}_i(u,v)=
\begin{cases}
\max\limits_{\mathbf{p}\in \mathcal{B}_{u,v}} z(\mathbf{p}), & \mathcal{B}_{u,v}\neq \emptyset,\\
0, & \text{otherwise}.
\end{cases}
\label{eq:scan_context}
\end{equation}

We store each descriptor together with its associated sampling position, forming the database
\begin{equation}
\mathcal{DB}=\left\{(\mathbf{p}_i,\mathbf{I}_i)\right\}_{i=1}^{N_p},
\label{eq:descriptor_db}
\end{equation}
which serves as the retrieval space for online relocalization.

\begin{algorithm}[!t]
\caption{Global Matching and Pose Estimation}
\label{alg:loop_closure}
\begin{algorithmic}[1]
\renewcommand{\algorithmicrequire}{\textbf{Inputs:}}
\renewcommand{\algorithmicensure}{\textbf{Output:}}

\REQUIRE Descriptor database $\mathcal{DB} = \{(\mathbf{p}_i, \mathbf{I}_i)\}_{i=1}^{N_p}$, \\
         \hspace{1.2em} Current online scans $\mathcal{P}_t^{\mathcal{L}}$ (sensor frame $\mathcal{L}$), \\
         \hspace{1.2em} Global map point cloud $\mathcal{G}$, \\
         \hspace{1.2em} Accumulation window $K_f$, candidate number $K_c$, \\
         \hspace{1.2em} Ray direction set $\mathcal{D}=\{\mathbf{d}_\ell\}_{\ell=1}^{K_r}$
\ENSURE Relocalized pose estimate $\mathbf{T}^* \in SE(3)$
\STATE $\mathcal{Q}_t \leftarrow \bigcup_{j=0}^{K_f-1} \mathcal{P}_{t-j}^{\mathcal{L}}$ 
\STATE $\mathcal{M}_t \leftarrow \text{Voxelize}(\mathcal{Q}_t; r)$
\STATE $\widehat{\mathcal{P}}_t^{\mathcal{L}} \leftarrow \text{RayCastFirstReturn}(\mathcal{M}_t; \mathcal{D}, R)$

\STATE $\mathbf{I}_t \leftarrow \text{SC\_GN}(\widehat{\mathcal{P}}_t^{\mathcal{L}}; N_r, N_s, L_{\max})$ 

\STATE $\{(\mathbf{p}_m, \hat{\psi}_m)\}_{m=1}^{K_c} \leftarrow \text{LOOP\_DE}(\mathbf{I}_t, \mathcal{DB}, K_c)$ 

\STATE $\tilde{\mathcal{Q}}_t \leftarrow \text{Down}(\mathcal{Q}_t), \ \tilde{\mathcal{G}} \leftarrow \text{Down}(\mathcal{G})$ 

\FOR{$m = 1$ \TO $K_c$}
    \STATE $\mathbf{T}_m^{(0)} \leftarrow [\mathbf{R}_z(\hat{\psi}_m) \mid \mathbf{p}_m]$
    \STATE $\mathbf{T}_m^* \leftarrow \text{GN\_ICP}(\tilde{\mathcal{Q}}_t, \tilde{\mathcal{G}}, \mathbf{T}_m^{(0)})$
    \STATE $\text{RMSE}_m \leftarrow FS_m(\tilde{\mathcal{Q}}_t, \tilde{\mathcal{G}}, \mathbf{T}_m^*)$  
    
    \IF{$\text{RMSE}_m \leq \tau$}
        \RETURN $\mathbf{T}^* \leftarrow \mathbf{T}_m^*$
    \ENDIF
\ENDFOR

\RETURN \textbf{fail}
\end{algorithmic}
\end{algorithm}

\subsection{Global Matching and Pose Estimate}
\label{subsec:multi_scan_accum_rep_cons}

The overall online global matching and pose-estimation procedure is summarized in Algorithm 1. Let $\mathcal{P}_t^{\mathcal{L}}$ denote the LiDAR point cloud at time $t$ expressed in the LiDAR frame $\mathcal{L}$. During relocalization, we assume the platform is approximately stationary and the accumulation window is sufficiently short, such that multiple scans can be treated as repeated observations under the same LiDAR pose. We therefore perform \emph{multi-scan accumulation} by directly stacking the most recent $K_f$ scans, thereby forming a more geometrically stable input for subsequent relocalization. The LiDAR runs at $20\,\mathrm{Hz}$; in all experiments we set $K_f=10$, yielding an accumulation horizon of $0.5\,\mathrm{s}$, which empirically provides adequate geometric support while maintaining the quasi-stationary assumption.
\begin{equation}
\mathcal{Q}_t \;=\; \bigcup_{j=0}^{K_f-1}\mathcal{P}_{t-j}^{\mathcal{L}} .
\label{eq:multi_scan_accum}
\end{equation}

To enforce \emph{representation consistency} between online global matching and the offline descriptor database (constructed from ray-casting-based synthetic scans), we voxelize $\mathcal{Q}_t$ into a local occupancy grid $\mathcal{M}_t$ and render a \emph{virtual scan} using the same first-return ray-casting procedure and LiDAR parameters (FoV, angular resolutions, and range limits) as in the offline stage. The rendered point set $\widehat{\mathcal{P}}_t^{\mathcal{L}}$ is then encoded into a Scan Context descriptor $\mathbf{I}_t$ using the same $(N_r, N_s, L_{\max})$ configuration as the database.

For global matching, we adopt the Scan Context-based place recognition method to query the descriptor database $\mathcal{DB}$ with $\mathbf{I}_t$. The retrieval module returns the top $K_c=5$ candidates, each providing a candidate location and an associated yaw estimate. Each candidate defines an initial pose hypothesis
\begin{equation}
\mathbf{T}^{(0)}_{m} \;=\;
\begin{bmatrix}
\mathbf{R}_z(\hat{\psi}_{m}) & \mathbf{p}_{m}\\
\mathbf{0}^\top & 1
\end{bmatrix},
\qquad m=1,\ldots,K_c,
\label{eq:init_pose_hypothesis}
\end{equation}
where $\mathbf{p}_m$ is the retrieved matching position and $\hat{\psi}_m$ is the estimated yaw angle.

Given an initial hypothesis $\mathbf{T}^{(0)}_m$, we refine the pose using GN-ICP between a downsampled source cloud $\widetilde{\mathcal{Q}}_t=\operatorname{Down}(\mathcal{Q}_t)$ and a downsampled global map cloud $\widetilde{\mathcal{G}}=\operatorname{Down}(\mathcal{G})$, with initialization $\mathbf{T}^{(0)}=\mathbf{T}^{(0)}_m$. In each iteration, correspondences are established (e.g., via nearest-neighbor search), and the pose is updated by minimizing the alignment error, following the standard GN-ICP procedure. We accept a refined hypothesis using the RMSE (PCL fitness score) reported by the PCL registration module. If the final RMSE is below a threshold $\tau$, the matching result is accepted and the final pose is set to $\mathbf{T}^{\ast}=\mathbf{T}^{\ast}_m$. Otherwise, we proceed to the next candidate $m\leftarrow m+1$ and repeat the GN-ICP refinement and RMSE verification until acceptance or until all $K_c$ candidates are exhausted. In practice, the best-ranked candidate typically provides a sufficiently accurate initialization for GN-ICP to converge; only a small fraction of degenerate cases trigger the fallback strategy.
\begin{figure*}
    \centering
    \includegraphics[width=1.0\linewidth]{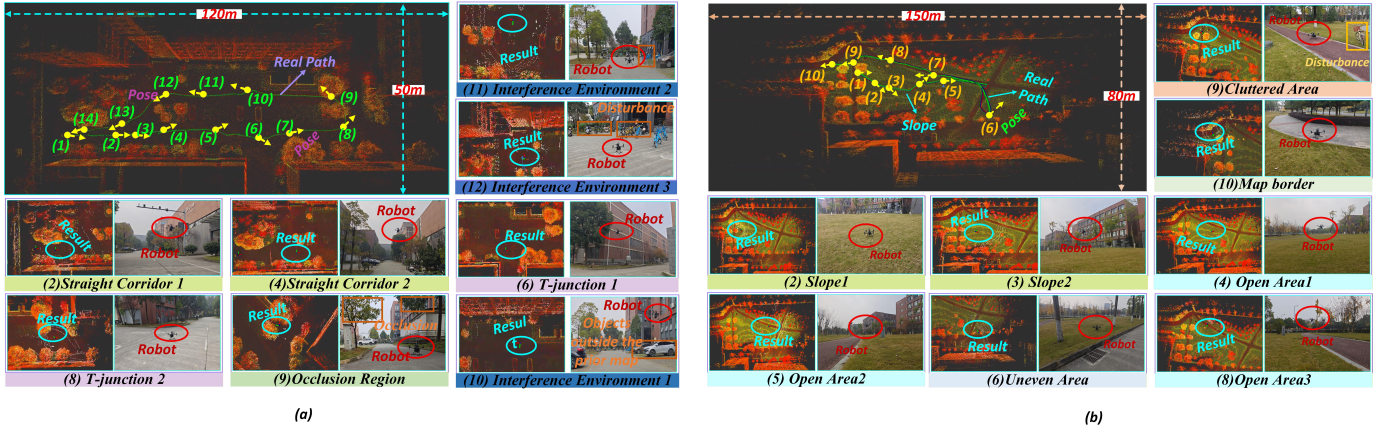}
    \caption{Two outdoor experimental scenarios and representative evaluation poses. (a) Wide corridor between buildings, approximately $120\,\mathrm{m}\times 50\,\mathrm{m}$. (b) Uneven grassland/slope environment, approximately $150\,\mathrm{m}\times 80\,\mathrm{m}$. The UAV flight trajectory is overlaid, and the numbered markers denote representative evaluation poses selected from typical regions (e.g., long corridors, cluttered areas, occluded areas, open spaces, T-junctions, and slopes). To emulate real-world disturbances, additional objects that are absent from the original map (e.g., bicycles and cars) are deliberately introduced (example locations are annotated in the figure).}
    \label{ex1}
\end{figure*}

\begin{figure}
    \centering
    \includegraphics[width=1.0\linewidth]{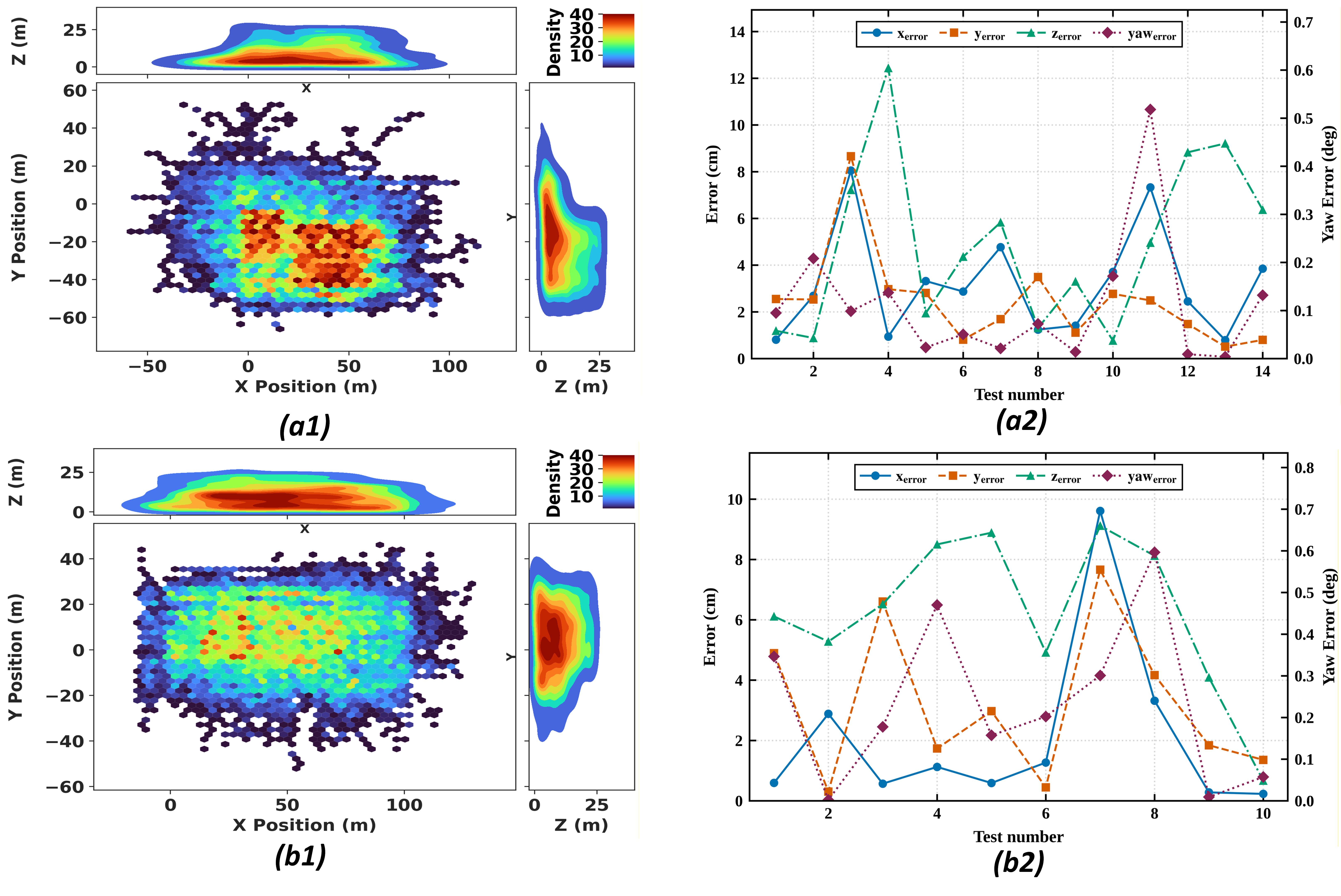}
    \caption{Visualization of constraint-aware feasible sampling and relocalization accuracy. (a1)(b1) Density visualization of the feasible sampled positions on the 3D occupancy grid map in the two scenarios: the planar (top-down) 2D density distribution is shown, and the marginal density of the $y\!-\!z$ projection is displayed along the boundaries; color indicates the local sample count/density. (a2)(b2) Mean 4-DoF errors in $x,y,z$, and $\mathrm{yaw}$ at representative poses (20 repeated trials per pose).}
    \label{fig:ex2}
\end{figure}
\begin{figure}
    \centering
    \includegraphics[width=1.0\linewidth]{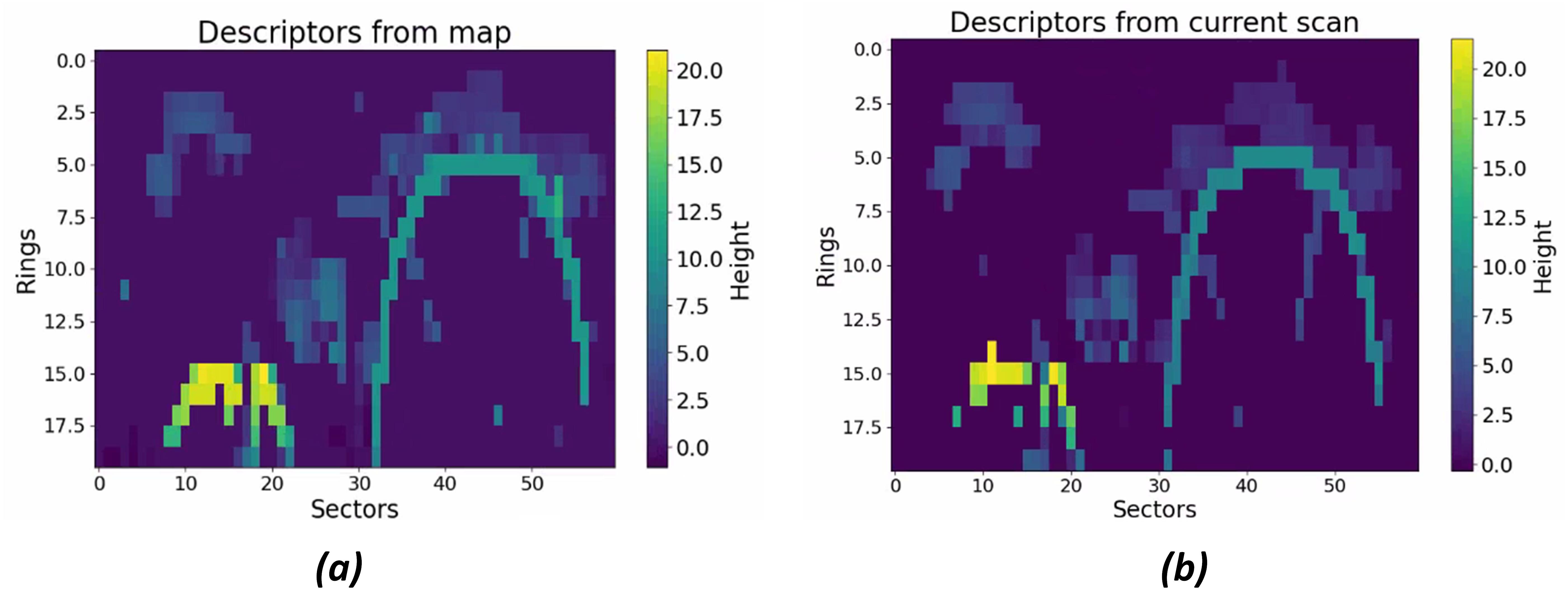}
    \caption{Qualitative comparison of Scan Context images produced by local-grid and global-map synthetic scans.
    (a) Scan Context generated from online accumulated LiDAR scans after voxelization into a local grid and sensor-centric first-return ray casting (no external pose input).
    (b) Scan Context generated by first-return ray casting on the global occupancy grid map at the relocalized position $\hat{\mathbf{p}}$.
    Both use $N_r=20$ rings and $N_s=60$ sectors with max-height encoding.}
    \label{fig:ex3}
\end{figure}

\section{Experiment And Discussion}    

We validate the proposed algorithm in two outdoor scenarios: a wide corridor between buildings and an uneven grassland environment, as illustrated in Fig.~\ref{ex1}(a) and (b), respectively. In both scenarios, comparative experiments are conducted under identical conditions using several alternative algorithms. To assess robustness and stability, the UAV is commanded to fly at different altitudes, execute varying turning/yaw angles, and traverse occluded regions. Scenario 1 (Fig.~\ref{ex1}(a)) is located between three buildings and covers approximately 120 m × 50 m, whereas Scenario 2 (Fig.~\ref{ex1}(b)) is situated on a slope with a size of approximately 150 m × 80 m. These environments encompass representative structures and challenges, including long corridors, cluttered areas, occlusions, open spaces, T-junctions, and slopes. To further emulate real-world disturbances, we deliberately introduce objects absent from the original map (e.g., bicycles and cars), such as those marked in Fig.~\ref{ex1}(a) (10)(11)(12) and Fig.~\ref{ex1}(b) (9). As shown in Fig.~\ref{fig:UAV}, our experimental platform consists of a UAV equipped with an MID360 LiDAR and an Intel NUC12 computer, running Ubuntu 20.04 and ROS1. Both the offline and online stages of relocalization are executed on this platform. The MID360 operates at 20 Hz with an effective range of 50 m. Since the LiDAR used in this work follows a non-repetitive scanning, a multi-frame accumulation strategy is employed to improve the geometric stability of the online query point clouds. Specifically, consecutive $K_f$ frames are merged into a single input point cloud, and we set $K_f=10$ throughout the experiments. We first build a global point-cloud map in both scenarios using our in-house SLAM system. Then, for each scenario, several representative poses are selected; when SLAM runs stably, the corresponding pose is recorded as the reference pose obtained from the stable SLAM trajectory for subsequent relocalization error evaluation, while LiDAR data around that location are simultaneously logged into rosbag files. For offline evaluation, the rosbags are replayed to export point-cloud sequences, and query point clouds are generated by accumulating $K_f=10$ frames. These accumulated query point clouds are used as the unified input for all compared methods, ensuring a fair and consistent experimental setup.

\newcommand{\hdrh}{\rule{0pt}{11pt}} 
\begin{table*}[t]
\centering
\footnotesize
\setlength{\tabcolsep}{4.0pt}
\renewcommand{\arraystretch}{1.15}
\caption{ Quantitative comparison in the building-corridor scene.}
\label{tab:scenario2_posewise}
\begin{tabular}{c|cccc|cccc|cccc|cccc}
\hline
\multirow{2}{*}{Pose\hdrh} &
\multicolumn{4}{c|}{Proposed\hdrh} &
\multicolumn{4}{c|}{GO-ICP\hdrh} &
\multicolumn{4}{c|}{FPFH-RANSAC+ICP\hdrh} &
\multicolumn{4}{c}{FPFH-FGR+ICP\hdrh} \\

\cline{2-17}
 & \makecell{\textbf{SR}\,(\%)\hdrh} 
 & \makecell{$\boldsymbol{\bar{t}}$\,(s)\hdrh} 
 & \makecell{$\boldsymbol{e_p}$\,(m)\hdrh} 
 & \makecell{$\boldsymbol{e_\psi}$\,($^\circ$)\hdrh}
 & \makecell{\textbf{SR}\,(\%)\hdrh} 
 & \makecell{$\boldsymbol{\bar{t}}$\,(s)\hdrh} 
 & \makecell{$\boldsymbol{e_p}$\,(m)\hdrh} 
 & \makecell{$\boldsymbol{e_\psi}$\,($^\circ$)\hdrh}
 & \makecell{\textbf{SR}\,(\%)\hdrh} 
 & \makecell{$\boldsymbol{\bar{t}}$\,(s)\hdrh} 
 & \makecell{$\boldsymbol{e_p}$\,(m)\hdrh} 
 & \makecell{$\boldsymbol{e_\psi}$\,($^\circ$)\hdrh}
 & \makecell{\textbf{SR}\,(\%)\hdrh} 
 & \makecell{$\boldsymbol{\bar{t}}$\,(s)\hdrh} 
 & \makecell{$\boldsymbol{e_p}$\,(m)\hdrh} 
 & \makecell{$\boldsymbol{e_\psi}$\,($^\circ$)\hdrh} \\
\hline
2  & \textbf{100} & \textbf{1.887} & \textbf{0.038} & \textbf{0.023} & 100 & 92.913  & 0.037 & 0.084 & 90  & 13.228 & 0.038 & 0.018 & 55 & 14.759 & 0.070 & 0.090 \\
4  & \textbf{90}  & \textbf{2.126} & \textbf{0.167} & \textbf{0.044} & 95 & 24.043  & 0.064 & 0.208 & 55  & 13.029 & 0.091 & 0.014 & 0  & /      & /     & /     \\
6  & \textbf{100} & \textbf{3.026} & \textbf{0.053} & \textbf{0.180} & 85  & 9.160   & 0.072 & 0.199 & 60  & 13.483 & 0.061 & 0.024 & 55 & 18.120 & 0.086 & 0.213 \\
8  & \textbf{100} & \textbf{1.621} & \textbf{0.039} & \textbf{0.071} & 100 & 39.322  & 0.083 & 0.096 & 60  & 13.477 & 0.062 & 0.010 & 65 & 16.384 & 0.465 & 0.107 \\
9  & \textbf{100} & \textbf{1.779} & \textbf{0.037} & \textbf{0.049} & 100  & 27.295  & 0.039 & 0.044 & 85  & 13.756 & 0.058 & 0.010 & 0  & /      & /     & /     \\
10 & \textbf{95}  & \textbf{1.175} & \textbf{0.047} & \textbf{0.132} & 95  & 61.834  & 0.093 & 0.066 & 65  & 13.214 & 0.085 & 0.026 & 25 & 22.924 & 1.115 & 1.658 \\
11 & \textbf{90}  & \textbf{2.153} & \textbf{0.092} & \textbf{0.072} & 90  & 11.318  & 0.094 & 0.255 & 70  & 18.239 & 0.088 & 0.674 & 10 & 20.763 & 1.269 & 0.219 \\
12 & \textbf{100}  & \textbf{1.775} & \textbf{0.093} & \textbf{0.009} & 100  & 26.538  & 0.085 & 0.033 & 35  & 11.203 & 0.103 & 0.008 & 0  & /      & /     & /     \\
\hline
\end{tabular}
\end{table*}

\begin{table*}[t]
\centering
\footnotesize
\setlength{\tabcolsep}{4.0pt}
\renewcommand{\arraystretch}{1.15}
\caption{Quantitative comparison in the grass-slope scene.}
\label{tab:scenario3_posewise}
\begin{tabular}{c|cccc|cccc|cccc|cccc}
\hline
\multirow{2}{*}{Pose\hdrh} &
\multicolumn{4}{c|}{Proposed\hdrh} &
\multicolumn{4}{c|}{GO-ICP\hdrh} &
\multicolumn{4}{c|}{FPFH-RANSAC+ICP\hdrh} &
\multicolumn{4}{c}{FPFH-FGR+ICP\hdrh} \\
\cline{2-17}
 & \makecell{\textbf{SR}\,(\%)\hdrh} 
 & \makecell{$\boldsymbol{\bar{t}}$\,(s)\hdrh} 
 & \makecell{$\boldsymbol{e_p}$\,(m)\hdrh} 
 & \makecell{$\boldsymbol{e_\psi}$\,($^\circ$)\hdrh}
 & \makecell{\textbf{SR}\,(\%)\hdrh} 
 & \makecell{$\boldsymbol{\bar{t}}$\,(s)\hdrh} 
 & \makecell{$\boldsymbol{e_p}$\,(m)\hdrh} 
 & \makecell{$\boldsymbol{e_\psi}$\,($^\circ$)\hdrh}
 & \makecell{\textbf{SR}\,(\%)\hdrh} 
 & \makecell{$\boldsymbol{\bar{t}}$\,(s)\hdrh} 
 & \makecell{$\boldsymbol{e_p}$\,(m)\hdrh} 
 & \makecell{$\boldsymbol{e_\psi}$\,($^\circ$)\hdrh}
 & \makecell{\textbf{SR}\,(\%)\hdrh} 
 & \makecell{$\boldsymbol{\bar{t}}$\,(s)\hdrh} 
 & \makecell{$\boldsymbol{e_p}$\,(m)\hdrh} 
 & \makecell{$\boldsymbol{e_\psi}$\,($^\circ$)\hdrh} \\
\hline
1  & \textbf{95}  & \textbf{1.640} & \textbf{0.049} & \textbf{0.347} & 0  & /      & /     & /     & 80  & 15.549 & 0.384 & 0.064 & 0  & /      & /     & /     \\
2  & \textbf{90}  & \textbf{2.071} & \textbf{0.066} & \textbf{0.002} & 75 & 12.452 & 0.082 & 0.350 & 75  & 21.670 & 0.062 & 0.020 & 25 & 20.351 & 0.491 & 9.688 \\
3  & \textbf{100} & \textbf{1.966} & \textbf{0.060} & \textbf{0.178} & 35 & 6.732  & 0.053 & 0.135 & 100 & 33.653 & 0.203 & 0.190 & 40 & 30.014 & 0.248 & 3.400 \\
4  & \textbf{100} & \textbf{1.751} & \textbf{0.087} & \textbf{0.470} & 75 & 9.053  & 0.126 & 0.661 & 90  & 39.682 & 0.153 & 0.049 & 30 & 36.527 & 1.392 & 0.098 \\
5  & \textbf{90}  & \textbf{2.055} & \textbf{0.094} & \textbf{0.157} & 80 & 8.995  & 0.107 & 0.289 & 90  & 38.000 & 0.194 & 0.218 & 10 & 36.187 & 3.514 & 4.770 \\
6  & \textbf{70}  & \textbf{2.426} & \textbf{0.092} & \textbf{0.065} & 0  & /      & /     & /     & 10  & 33.750 & 0.121 & 0.087 & 0  & /      & /     & /     \\
8  & \textbf{100} & \textbf{1.630} & \textbf{0.051} & \textbf{0.218} & 70 & 11.519 & 0.051 & 0.047 & 80  & 29.449 & 0.109 & 0.069 & 25 & 28.637 & 0.319 & 5.490 \\
9  & \textbf{90}  & \textbf{1.887} & \textbf{0.116} & \textbf{0.417} & 95 & 7.802  & 0.095 & 0.186 & 95  & 30.315 & 0.163 & 0.187 & 15 & 29.971 & 0.735 & 8.735 \\
10 & \textbf{95}  & \textbf{1.457} & \textbf{0.073} & \textbf{0.308} & 0  & /      & /     & /     & 70  & 33.806 & 0.079 & 0.036 & 0  & /      & /     & /     \\
\hline
\end{tabular}
\end{table*}

To visually evaluate the sampling outcomes of the proposed constraint-aware feasible sampling strategy on the 3D occupancy grid map, we visualize the density of the feasible sampled positions in the two experimental scenarios (Fig.~\ref{fig:ex2}(a1) and Fig.~\ref{fig:ex2}(b1)). Specifically, we show the 2D density distribution on the planar projection and display the marginal densities along the $y$--$z$ projection on the boundaries (where color indicates the local sample count/density). The results demonstrate that the samples exhibit higher density and more continuous coverage in open, traversable areas of the map; in contrast, sampling becomes markedly sparse near obstacle surfaces or close to map boundaries, reflecting the suppression of unsafe locations or regions with insufficient map information. Fig. \ref{fig:ex3} illustrates the qualitative difference between descriptors from map-based ray casting and current LiDAR scan; this level of similarity is sufficient for global matching in relocalization.

Considering that the aerial platform in this task mainly involves translation and heading, we evaluate only the 4-DoF errors in $x,y,z$ and yaw. For each relocalization result $\hat{\mathbf{T}}$ and the reference pose $\mathbf{T}_{ref}$, the positional error $d_{err}$ is defined as the Euclidean distance between their translations, while the heading error $\psi_{err}$ is defined as the absolute difference in yaw. For each pose, we repeat the experiment 20 times; the success rate (SR) is computed as the ratio of trials that satisfy the success criterion, and the efficiency is measured by the average runtime $t$ (from point-cloud input to pose output). In addition, we report the mean positional error $e_p$ (m) and the mean heading error $e_\psi$ ($^\circ$) to quantify accuracy. A relocalization trial is regarded as successful if $d_{err}\le 4\,\mathrm{m}$ and $\psi_{err}\le 20^\circ$, which is sufficient for initializing subsequent 3D SLAM/navigation. As shown in Fig.~7(a2) and Fig.~7(b2), we further present the average $x,y,z,$ yaw errors in the two experimental scenarios.

As shown in Fig.~\ref{ex1}, we select representative poses in each scenario, covering long straight corridors, T-junctions, occlusion/disturbance regions, and map-boundary areas; the quantitative comparisons are reported in Tables~\ref{tab:scenario2_posewise} and~\ref{tab:scenario3_posewise}. The results demonstrate strong cross-scenario robustness of the proposed method: under occlusion (Pose~9) and disturbance settings (Poses~10--12), it consistently achieves a success rate (SR) above 85\%, whereas FPFH-FGR+ICP is prone to complete failure in such regions (SR=0) and GO-ICP incurs a significant runtime increase up to 61.834\,s. In challenging boundary-related cases, including the map boundary (Pose~10, Table~II) and uneven terrain (Pose~6, Table~II), conventional methods such as GO-ICP and FPFH-FGR+ICP degrade severely (SR=0), while the proposed method still maintains an SR of 70\%--95\%. Moreover, the proposed method delivers stable outputs with second-level latency across all challenging poses (average $<3\,\mathrm{s}$); for successful relocalization trials, the mean position error $e_p$ remains below $0.15\,\mathrm{m}$ and the mean heading error $e_\psi$ remains within $1^\circ$. Overall, the proposed approach outperforms the baselines in success rate, real-time efficiency, and accuracy, making it well suited for online relocalization that requires both timeliness and precision.

\section{Conclusion And Future Work}

To address the kidnapped robot problem in GNSS-denied environments, this paper proposes a LiDAR-based global
3D relocalization method. Compared with existing feature-matching approaches, the core contribution of this work lies in the proposal of a constraint-aware 3D uniform sampling strategy and a ray-casting-based place recognition mechanism. The system's effectiveness was verified in two typical outdoor scenarios: the proposed method achieves a high success rate of 90\%--100\% while maintaining high positioning accuracy ($e_p < 0.1$~m). Furthermore, compared to the traditional GO-ICP method, the proposed approach achieves an order-of-magnitude improvement in computational efficiency (reducing runtime from tens of seconds to the second level), demonstrating significant potential for practical engineering applications.

Future work will focus on the following aspects: First, to address large roll/pitch variations often encountered by UAVs or legged robots in extreme terrain, we plan to improve existing descriptors and incorporate ground geometric constraints to achieve robust initialization under full 6-DoF. Second, regarding multi-modal fusion, we aim to integrate visual information (e.g., from onboard cameras) into the current framework. By constructing a joint visual-LiDAR representation, we seek to further enhance the system's robustness in environments with sparse geometric features or structural degeneration.

\bibliographystyle{IEEEtran}
\bibliography{reference}

\end{document}